\documentstyle{llncs}

\newcommand{\mytuple}[1]{\langle #1 \rangle}

\begin{document}
\title{Bounds on sample size for policy evaluation in Markov environments}

\author{Leonid Peshkin and Sayan Mukherjee}

\institute{MIT Artificial Intelligence Laboratory\\
 545 Technology Square\\
 Cambridge, MA 02139\\
 {\tt \{pesha,sayan\}@ai.mit.edu}\\
}

\maketitle

\begin{abstract}
Reinforcement learning means finding the optimal course of action
in Markovian environments without knowledge of the environment's dynamics.
Stochastic optimization algorithms used in the field rely on estimates of the
value of a policy. Typically, the value of a policy is estimated from results
of simulating that very policy in the environment. This approach requires a
large amount of simulation as different points in the policy space are
considered. In this paper, we develop value estimators that utilize data
gathered when using one policy to estimate the value of using another policy,
resulting in much more data-efficient algorithms. We consider the question of
accumulating a sufficient experience and give PAC-style bounds.
\end{abstract}

\section{Introduction}
\label{intro}

Research in reinforcement learning focuses on designing algorithms for an
agent interacting with an environment, to adjust its behavior in such a way
as to optimize a long-term return. This means searching for an optimal
behavior in a class of behaviors. Success of learning algorithms
therefore depends both on the richness of information about various behaviors
and how effectively it is used. While the latter aspect has been
given a lot of attention, the former aspect has not been addressed
scrupulously. This work is the attempt to adapt solutions developed for
similar problems in the field of statistical learning theory. 

The motivation for this work comes from the fact that, in reality, the
process of interaction between the learning agent and the environment is
costly in terms of time, money or both. Therefore, it is important to
carefully allocate available interactions, to use all available information
efficiently and to have an estimate of how informative the experience overall
is with respect to the class of possible behaviors. The interaction between
agent and environment is modeled by a Markov decision process
({\sc mdp})~\cite{Kaelbling96,Puterman94}. The learning system does not know the
correct behavior, or the true model of the environment it interacts with.
Given the sensation of the environment state as an input, the agent
chooses the action according to some rule, often called a {\em policy}. This
action constitutes the output. The effectiveness of the action taken and its
effect on the environment is communicated to the agent through a scalar value
({\em reinforcement signal}).

The environment undergoes some transformation---changes the current state
into a new state. A few important assumptions about the environment are made.
In particular, the so-called Markov property is assumed: given the current
state and action, the next state is independent of the rest of the history of
states and actions. Another assumption is a {\em non-deterministic}
environment, which means that taking the same action in the same state could
lead to a different next state and generate a different payoff signal. It is
an objective of the agent to find a behavior which optimizes some long-run
measure of payoff, called {\em return}.

There are many efficient algorithms for the case when the agent has perfect
information about the environment. An optimal policy is described by mapping
the last observation into an action and can be computed in polynomial time in
the size of the state and action spaces and the effective time
horizon~\cite{Bellman57}. However in many cases the environment
state is described by a vector of several variables, which makes the
environment state size exponential in the number of variables. Also, under
more realistic assumptions, when a model of environment dynamics is unknown
and the environment's state is not observable, many problems arise. The
optimal policy could potentially depend on the whole history of interactions
and for the undiscounted finite horizon case computing it is {\sc
pspace}-complete~\cite{Papadimitriou87}.
In realistic settings, the class of policies is restricted and even among the
restricted set of policies, the absolute best policy is not expected to be
found due to the difficulty of solving a global multi-variate optimization
problem. Rather, the only option is to explore different approaches to
finding near-optimal solutions among local optima.

The issue of finding a near-optimal policy from a given class of policies is
analogous to a similar issue in supervised learning. There we are looking for
a near-optimal hypothesis from a given class of hypotheses~\cite{Vapnik98}.
However, there are crucial differences in these two settings. In supervised
learning we assume that there is some target function, that labels the
examples, and some distribution that generates examples. A crucial property
is that the distribution is the same for all the hypotheses. This implies
both that the same set of samples can be evaluated on any hypothesis, and
that the observed error is a good estimate of the true error.

On the other hand, there is no fixed distribution generating experiences in
reinforcement learning. Each policy induces a different distribution over
experiences. The choice of a policy defines both a ``hypothesis'' and a
distribution. This raises the question of how one re-uses the experience
obtained while following one policy to learn about another. The other policy
might generate a very different set of samples (experiences), and in the
extreme case the support of the two distributions might be disjoint.

In the pioneering work by Kearns et al.~\cite{KearnsNIPS99}, the issue of
generating enough information to determine the near-best policy is
considered. Using a random policy (selecting actions uniformly at random),
they generate a set of history trees. This information is used to define
estimates that uniformly converge to the true values. However, this work
relies on having a generative model of the environment, which allows
simulation of a reset of the environment to any state and execute any action
to sample an immediate reward. Also, the reuse of information is partial---an
estimate of a policy value is built only on a subset of experiences,
``consistent'' with the estimated policy.

Mansour~\cite{Mansour99} has addressed the issue of computational complexity
in the setting of Kearns et al.~\cite{KearnsNIPS99} by establishing a
connection between {\em mistake bounded algorithms} (adversarial on-line
model~\cite{Littlestone88}) and computing a near-best policy from a given
class with respect to a finite-horizon return. Access to an algorithm that
learns the policy class with some maximal permissible number of mistakes is
assumed. This algorithm is used to generate ``informative'' histories in the
{\sc pomdp}, following various policies in the class, and determine a
near-optimal policy. In this setting a few improvements in bounds are made.

In this work we present a way of reusing all of the accumulated experience
without having access to a generative model of the environment. We make use of
the technique known as ``importance sampling''~\cite{Rubinstein81} or
``likelihood ratio estimation''~\cite{Glynn89} to different communities. We
discuss properties of different estimators and provide bounds for the uniform
convergence of estimates on the policy class. We suggest a way of using these
bounds to select among candidate classes of policies with various complexity,
similar to structural risk minimization~\cite{Vapnik98}.

The rest of this paper is organized as follows. 
Section~\ref{optim} discusses reinforcement learning as a stochastic optimization problem. 
In Section~\ref{notation} we define our notation. 
Section~\ref{sampling} presents the necessary background in sampling theory
and presents our way of estimating the value of policies. 
The algorithm and {\sc pac}-style bounds are given in Section~\ref{algorithm}.

\section{Reinforcement Learning as Stochastic Optimization}
\label{optim}
There are various approaches to solving {\sc rl}~problems. {\em Value
search} algorithms find the optimal policy by using
dynamic-programming methods to compute the value function---utility of
taking a particular action in a particular world state---then
deducing the optimal policy from the value function. {\em Policy
search} algorithms (e.g. {\sc reinforce}~\cite{Williams92}) work directly in
the policy class, trying to maximize the expected reward without the
help of Bellman's optimality principle. 

Policy search methods rely on estimating the value of the policy (or the
gradient of the value) at various points in a policy class and attempt to
solve the {\em optimization issue}. In this paper we ignore the optimization
issue and concentrate on the {\em estimation issue}---how much and what kind
of experience one needs to generate in order to be able to construct
uniformly good value estimators over the whole policy class. In particular we
would like to know what the relation is between the number of sample
experiences and the confidence of value estimates across the policy class.

Two different approaches to optimization can be taken. One involves using an
algorithm, driven by newly generated policy value (or gradient thereof)
estimates at each iteration to update the hypothesis about the optimal policy
after each interaction (or few interactions) with the environment. We will
call this {\em on-line} optimization. Another is to postpone optimization
until all possible interaction with the environment is exhausted, and combine
all information available in order to estimate ({\em off-line}) the whole
``value surface''.

In this paper we are not concerned with the question of optimization. We
concentrate on the second case with the goal of building a module that
contains a non-parametric model of optimization surface. Given an arbitrary
policy such a module outputs an estimate of its value, as if the policy was
tried out in the environment. Once such module is built and guarantees on 
good estimates of policy value are obtained across the policy class, we may
use our favorite optimization algorithm. Gradient descent methods, in
particular {\sc reinforce}~\cite{Williams92,WIlliams86} have been used recently in
conjunction with policy classes constrained in various ways, e.g., with
external memory~\cite{Peshkin99}, finite state
controllers~\cite{MeuleauUAI99} and in multi-agent settings~\cite{Peshkin00}.
Furthermore, the idea of using importance sampling in the reinforcement
learning has been explored~\cite{MeuleauTR00,PrecupICML00}. However only
on-line optimization was considered.

One realistic off-line scenario in reinforcement learning is when the data
processing and optimization (learning) module is separated (physically) from
the data acquisition module (agent). Say we have an ultra-light micro-sensor
connected to a central computer . The agent then has to be instructed
initially how to behave when given a chance to interact with the environment
for a limited number of times, then bring/transmit the collected data back.
Naturally during such limited interaction only a few possible behaviors can
be tried out. It is extremely important to be able to generalize from this
experience in order to make a judgment about the quality of behaviors which
were not tried out. This is possible when some kind of similarity measure in
the policy class can be established. If the difference between the
values of two policies could be estimated, we could estimate the value of one
policy based on experience with the other.

\section{Background and Notation}
\label{notation}
\paragraph{MDP}
\label{mdps}
The class of problems described above can be modeled as Markov decision processes ({\sc mdp}s).
An {\sc mdp}~ is a 4-tuple $\mytuple{ S,A,T,R}$, where: $S$ is the set of states;
$A$ is the set of actions; $T\!:\!S\!\times\!{\cal A}\!\rightarrow\!{\cal
P}(S)$ is a mapping from states of the environment and actions of the agent
to probability distributions\footnote{Let ${\cal P}(\Omega)$ denote the set
of probability distributions defined on some space $\Omega$.} over states of
the environment; and \mbox{$R\!:\!S\!\times\!A \rightarrow R$} is the payoff
function ({\em reinforcement}), mapping states of the environment and actions
of the agent to immediate reward.

\paragraph{POMDP}
The more complex case is when the agent is no longer able to reliably
determine which state of the {\sc mdp}~ it is currently in. The process of
generating an observation is modeled by an observation function $B(s(t))$.
The resulting model is a {\em partially observable Markov decision process}
({\sc pomdp}). 
Formally, a {\sc pomdp}~ is defined as a tuple $\mytuple{S, O, A, B, T, R}$ where:
$S$ is the set of states;
$O$ is the set of observations;
$A$ is the set of actions;
$B$ is the observation function $B\!: S \rightarrow{\cal P}(O)$;
$T\!:\!S\!\times\!{\cal A}\!\rightarrow\!{\cal P}(S)$ is a mapping
from states of the environment and actions of the agent to probability
distributions over states of the environment;
\mbox{$R\!:\!S\!\times\!A \rightarrow {\cal R}$} is the payoff
function, mapping states of the environment and actions of the agent to
immediate reward.
In a {\sc pomdp}, at each time step: an agent observes $o(t)$
corresponding to $B(s(t))$ and performs an action $a(t)$ according to its
policy, inducing a state transition of the environment; then receives the
reward $r(t)$. We assume that the rewards $r(s,a)$ are bounded by $r_{max}$
for any $s$ and $a$.

\paragraph{History}
We denote by $H_t$ the set of all possible experience sequences of length
$t$: \mbox{$H_t = \left\{
\mytuple{o(1),a(1),r(1),\ldots,o(t),a(t),r(t),o(t+1)}\right\}$}, where
\mbox{$o(t) \in O$} is the observation of agent at time $t$; \mbox{$a(t) \in
A$} is the action the agent has chosen to take at time $t$; and \mbox{$r(t)
\in {\cal R}$} is the reward received by agent at time $t$. In order to
specify that some element is a part of the history $h$ at time $\tau$, we
write, for example, $r(\tau,h)$ and $a(\tau,h)$ for the $\tau^{th}$ reward
and action in the \mbox{history} $h$.

\paragraph{Policy}
Generally speaking, in a {\sc pomdp}, a policy $\pi$ is a rule specifying the
action to perform at each time step as a function of the whole previous
history: \mbox{$\pi: H \rightarrow {\cal P}(A)$}. {\em Policy class} $\Pi$
is any set of policies. We assume that the probability of the elementary
event is bounded away from zero: \mbox{$0 \le \underline{c} \le \Pr(a|h,\pi)$}, for any
$a\in A$, $h\in H$ and $\pi\in\Pi$.

\paragraph{Return}
A history $h$ includes several immediate rewards $\mytuple{r(1) \ldots r(i)
\ldots }$, that can be combined to form a return $R(h)$. In this paper we
focus on returns which may be computed (or approximated) using the first $T$
steps, and are bounded in absolute value by $R_{max}$. This includes two
well-studied return functions---the undiscounted finite horizon return and
the discounted infinite-horizon return. The first is \mbox{$R(h)
=\sum_{t=1}^T r(t,h)$}, where $T$ is the finite-horizon length. In this case
\mbox{$R_{max} = T r_{max}$}. The second is the discounted infinite horizon
return~\cite{Puterman94} \mbox{$R(h) = \sum_{t = 0}^\infty \gamma^t
r(t,h)$}, with a geometric discounting by the factor \mbox{$\gamma \in (0;1)$}.
In this case we can approximate $R$ using the first $T_\epsilon = \log_\gamma
\frac{\epsilon}{R_{max}}$ immediate rewards. Using $T_\epsilon$ steps we can
approximate $R$ within $\epsilon$ since \mbox{$R_{max} = \frac{r_{max}}{1
- \gamma}$} and \mbox{$\sum_{t = 0}^\infty \gamma^t r(t) - \sum_{t =
0}^{T_\epsilon} \gamma^t r(t) < \epsilon$}. It is important to approximate
the return in $T$ steps, since the length of the horizon is a parameter in our bounds.

\paragraph{Value}
Any policy \mbox{$\pi \in \Pi$} defines a conditional distribution
$\Pr(h|\pi)$ on the class of all histories $H$. The value of policy $\pi$ is
the expected return according to the probability induced by this policy on
histories space:
$$
V(\pi) = {\rm E}_{\pi} \left[ R(h) \right] = \sum_{h \in H} 
\left[ R(h) \Pr(h|\pi) \right],  
$$
where for brevity we introduced notation ${\rm E}_{\pi}$ for ${\rm
E}_{\Pr(h|\pi)}$. It is an {\em objective} of the agent to find a policy
$\pi^*$ with optimal value: $\pi^* = {\rm argmax}_{\pi} V(\pi)$. We assume
that policy value is bounded by $V_{max}$. That means of course that returns
are also bounded by $V_{max}$ since value is a weighted sum of returns.

\section{Sampling}
\label{sampling}

For the sake of clarity we are introducing concepts from sampling theory
using functions and notation for relevant reinforcement learning concepts.
Rubinstein~\cite{Rubinstein81} provides a good overview of this material.

\paragraph{``Crude'' sampling}
If we need to estimate the value $V(\pi)$ of policy $\pi$, 
from independent, identically distributed (i.i.d.) samples induced by this
policy, after taking $N$ samples \mbox{$h_i, i \in (1..N)$} we have:
$$
\hat V(\pi) = \frac{1}{N} \sum_i R(h_i)\;\;.
$$
The expected value of this estimator
is $V(\pi)$ and it has variance ${\rm Var}\!\left[ \hat V(\pi) \right]$: 
$$
\frac{1}{N} \sum_{h \in H} R(h)^2 \Pr(h|\pi) - \frac{1}{N} \left[ \sum_{h \in H} R(h)
\Pr(h|\pi) \right] ^2 \!=\!  \frac{1}{N} {\rm E}_{\pi}\!\left[ R(h)^2 \right] - \frac{1}{N} V^2(\pi)\;\;.
$$

\paragraph{Indirect sampling}
Imagine now that for some reason we are unable to sample from the policy
$\pi$ directly, but instead we can sample from another policy $\pi'$. The
intuition is that if we knew how ``similar'' those two policies were to one
another, we could use samples drawn according to the distribution $\pi'$ and
make an adjustment according to the similarity of the policies. Formally we
have:
\begin{eqnarray*}
V(\pi) =& \sum_{h \in H} R(h_i) \Pr(h|\pi) = \sum_{h \in H} R(h_i) 
        \frac{\Pr(h|\pi')}{\Pr(h|\pi')} \Pr(h|\pi)\\
  =& \sum_{h \in H} R(h_i) \frac{\Pr(h|\pi)}{\Pr(h|\pi')} \Pr(h|\pi') = 
      {\rm E}_{\pi'} \left[ R(h_i) \frac{\Pr(h|\pi)}{\Pr(h|\pi')}\right]\;\;,
\end{eqnarray*}
where an agent might not be (and most often is not) able to calculate
$\Pr(h|\pi)$. 

\begin{lemma}
\label{lr}
It is possible to calculate $\frac{\Pr(h|\pi)}{\Pr(h|{\pi'})}$ for any $\pi,
\pi' \in \Pi$ and $h\in H$. 
\end{lemma}
\begin{proof} 
The Markov assumption in {\sc pomdp}s warrants that
\begin{eqnarray*}
\Pr(h|\pi) &=& \Pr(s(0)) \prod_{t=1}^T \Pr(o(t)|s(t)) \Pr(a(t)|o(t),\pi) \Pr(s(t+1)|s(t),a(t)) \\
   &=& \left[\Pr(s(0)) \prod_{t=1}^T \Pr(o(t)|s(t))\Pr(s(t+1)|s(t),a(t))\right]
      \left[ \prod_{t=1}^T \Pr(a(t)|o(t),\pi) \right] \\
   &=& \Pr(h_e) \Pr(h_a|\pi)\,\,.
\end{eqnarray*}
$\Pr(h_e)$ is the probability of the part of the history, dependent on the
environment, that is unknown to the agent and can be only sampled.
$\Pr(h_a|\pi)$ is the probability of the part of the history, dependent on
the agent, that is known to the agent and can be computed (and
differentiated). Therefore we can compute $$\frac{\Pr(h|\pi)} 
{\Pr(h|{\pi'})} = \frac{\Pr(h_e)\Pr(h_a|\pi)}{\Pr(h_e)\Pr(h_a|{\pi'})} =
\frac{\Pr(h_a|\pi)}{\Pr(h_a|{\pi'})}\;\;.$$ \qed
\end{proof}

We can now construct an indirect estimator $\hat V_{\pi'}(\pi)$ from i.i.d. samples
$h_i, i \in (1..N)$ according to the distribution $\Pr(h|\pi')$:
\begin{equation}
\label{eq_IS}
\hat V_{\pi'}(\pi) = \frac{1}{N} \sum_i R(h_i) w_{\pi}(h_i,\pi')\;\;,
\end{equation}
where for convenience, we denote the fraction $\frac{\Pr(h|\pi)}{\Pr(h|{\pi'})}$ by $w_{\pi}(h,\pi')$.   
This is an unbiased estimator of $V(\pi)$ with variance
\begin{equation}
\label{eq_ISVar}
\begin{array}{rcl}
{\rm Var}\left[ \hat V_{\pi'}(\pi) \right] &=&  \frac{1}{N} \left\{ \sum_{h \in H}
\left( R(h) w_{\pi}(h,\pi') \right) ^2 \Pr(h|\pi') - {V(\pi)}^2 \right\} \\
&=& \frac{1}{N} \left\{ \sum_{h \in H}
\frac{({R(h)\Pr(h|\pi))}^2}{\Pr(h|\pi')} - {V(\pi)}^2 \right\}  \\
&=& \frac{1}{N} {\rm E}_{\pi} \left[ R(h)^2 w_{\pi}(h,\pi') \right] -
   \frac{1}{N} {V(\pi)}^2 \;\;.
\end{array}
\end{equation}

This estimator $\hat V_{\pi'}(\pi)$ is usually called in
statistics~\cite{Rubinstein81} an {\em importance sampling} ({\sc is}) estimator
because the probability $\Pr(h|\pi')$ is chosen to emphasize parts of the
sampled space that are important in estimating $V$. The technique of {\sc is}~was
originally designed to increase the accuracy of Monte Carlo estimates by
reducing their variance~\cite{Rubinstein81}. Variance reduction is always a
result of exploiting some knowledge about the estimated quantity.

\paragraph{Optimal sampling policy}
It can be shown~\cite{Kahn53}, for example by optimizing the
expression~\ref{eq_ISVar} with Lagrange multipliers, that the
optimal sampling distribution is $\Pr(h|\pi') = \frac{R(h)\Pr(h|\pi)
}{V(\pi)}$, which gives an estimator with {\em zero}
variance. Not surprisingly this distribution can not be used, since it
depends on prior knowledge of a model of the environment (transition
probabilities, reward function), which contradicts our assumptions,
and on the value of
the policy which is what we need to calculate. However all is not 
lost. There are techniques which approximate the optimal distribution,
by changing the sampling distribution during the trial, while keeping
the resulting estimates unbiased via reweighting of samples, called
"adaptive importance sampling" and "effective importance sampling"
(see, for example, ~\cite{oh89,Zhou98,Ortiz00}).
In the absence of any information about $R(h)$ or estimated policy, the
optimal sampling policy is the one which selects actions uniformly at random:
$\Pr(a|h)=\frac{1}{2}$. For the horizon $T$, this gives us the upper bound
which we denote $\eta$: 
\begin{equation}
\label{eq_w}
w_{\pi}(h,\pi') \leq 2^T(1-\underline{c})^T = \eta \;\;.
\end{equation}

\begin{remark}
One interesting observation is that it is possible to get a better estimate
of $V(\pi)$ while following another policy $\pi'$. Here is an illustrative
example: imagine that reward function $R(h)$ is such that it is {\em zero}
for all histories in some sub-space $H_0$ of history space $H$. At the same
time policy $\pi$, which we are trying to estimate spends almost all the time
there, in $H_0$. If we follow $\pi$ in our exploration, we are wasting
samples/time! In this case, we can really call what happens {\em "importance
sampling"}, unlike usually when it is just "reweighting", not connected to
"importance" {\it per se}. That is why we advocate using the name {\em
``likelihood ratio''} rather than {\em``importance sampling''}.
\end{remark}

\begin{remark}
So far, we talked about using a single policy to collect all samples for
estimation. We also made an assumption that all considered distributions
have equal support. In other words, we assumed that any history has a non-zero
probability to be induced by any policy. 
Obviously it could be beneficial to execute a few different sampling
policies, which might have disjoint or overlapping support. There is
literature on this so-called stratification sampling
technique~\cite{Rubinstein81}. Here we just mention that it is possible to extend
our analysis by introducing a prior probability on choosing a policy
out of a set of sampling policies, then executing this sampling
policy. Our sampling probability will become: \mbox{$\Pr(h) = \Pr(\pi')\Pr(h|\pi')$}.
\end{remark}

\section{Algorithm and Bounds}
\label{algorithm}

Table~1 presents the computational procedure for estimating the
value of any policy from the policy class off-line. The sampling stage
consists of accumulating histories $h_i$, $i \in [1..N]$ induced by a
sampling policy $\pi'$ and calculating returns on these histories $R(h_i)$.
After the first stage is done, the procedure can simulate the interaction
with the environment for any policy search algorithm, by returning an
estimate for arbitrary policy.

\begin{table}[ht]
{\footnotesize \caption{Policy evaluation}
\rule{12cm}{1pt}
\begin{tabbing}
111\=111\=11111111111111111111111111111111111111\=1\=111\= \kill
  \> {\bf Sampling stage:} \\
  \> \> Chose a sampling policy $\pi'$; \\
  \> \> Accumulate the set of histories $h_i$, $i \in [1..N]$ induced by  $\pi'$; \\
  \> \> Calculate the set of returns $R(h_i)$, $i \in [1..N]$; \\
  \> {\bf Estimation stage}: \\
  \> \> Input: policy $\pi \in \Pi$ \\
  \> \> Calculate $w_{\pi}(h_i,\pi')$ for $i \in [1..N]$;  \\
  \> \> Output: estimate $\hat V(\pi)$ according to equation~\ref{eq_IS}:
$\frac{1}{N} \sum_i R(h_i) w_{\pi}(h_i,\pi')$
\end{tabbing}
}
\rule{12cm}{1pt}
\end{table}

\subsection{Sample Complexity}

We first compute deviation bounds for the {\sc is}~ estimator for a
single policy from its expectation using Bernstein's inequality:
\begin{theorem}(Bernstein~\cite{Bernstein46})
\label{th_Bernstein}
Let $\xi_1,\xi_2, \ldots$ be independent random variables with identical mean
${\rm E}\xi$, bounded by some constant  $|\xi_i| \leq a$, $a > 0 $. 
Also let ${\rm Var}(M_N)= {\rm E}\xi_1^2 + \ldots + {\rm E}\xi_N^2 \leq L$.
Then the partial sums $M_N = \xi_1 + \ldots + \xi_N$ obey the following
inequality for all $\epsilon > 0$:
$$
\Pr \left( \left| \frac{1}{N} M_N - E \xi \right| > \epsilon \right) \leq
2\exp{\left({-\frac{1}{2}\frac{\epsilon^2 N}{L+a\epsilon}}\right)}.
$$
\end{theorem}  

\begin{lemma}
\label{th_dd}
With probability $(1-\delta)$ the following holds true.
The estimated value $\hat V(\pi)$ based on $N$ samples is close to the true
value $V(\pi)$ for some policy $\pi$ : 
\begin{eqnarray*}
\left| V(\pi) - \hat V(\pi)\right| \le \frac{V_{max}}{N} \left
( \log(1/\delta) \eta +\sqrt{2 \log(1/\delta) (\eta-1) + \log(1/\delta)^2 \eta^2} \right).
\end{eqnarray*}
\end{lemma}
\begin{proof}
In our setup, 
$\xi_i = R(h_i) w_{\pi}(h_i,\pi')$, and ${\rm E}\xi = 
{\rm E}_{\pi'} \left[ R(h_i) w_{\pi}(h_i,\pi')\right] = 
{\rm E}_{\pi} \left[  R(h_i) \right] = V(\pi)$; 
and $a = V_{max} \eta$ by equation~\ref{eq_w}.
According to equation~\ref{eq_ISVar} $L={\rm Var}(M_N) 
={\rm Var}\hat V_{\pi'}(\pi) \le \frac{V^2_{max}}{N}(\eta -1)$. 
So we can use Bernstein's inequality and we get the
following deviation bound for a policy $\pi$:
\begin{equation}
\label{our_Bern}
\Pr \left( \left| V(\pi) - \hat V(\pi)  \right| > \epsilon \right) \leq
2 \exp \left[{-\frac{1}{2}\frac{\epsilon^2 N}{ \frac{V^2_{max}(\eta-1)}{N}
+ V_{max} \eta \epsilon}}\right]
= \delta,
\end{equation}

After solving for $\epsilon$, we get the statement of Lemma~\ref{th_dd}. \qed
\end{proof}

Note that this result is for a single policy. We need a convergence result
{\em simultaneously} for all policies in the class $\Pi$. 
We proceed using classical uniform convergence results for covering numbers
as a measure of complexity. 

\begin{remark}
We use covering numbers (instead of VC dimension as Kearns et
al.~\cite{KearnsNIPS99}) both as a measure of the metric complexity of a
policy class in a union bound and as a parameter for bounding the likelihood
ratio. Another advantage is that metric entropy is a more refined measure of
capacity than VC dimension since the VC dimension is an upper bound on
the growth function which is an upper bound on the metric entropy~\cite{Vapnik98}.
\end{remark}

\begin{definition}
Let $\Pi$ be class of policies that form a metric space with
metric $D_{\infty}(\pi,\pi')$ and  $\varepsilon > 0$. The covering number
${\cal{N}}(\Pi,D,\varepsilon)$ is defined as the minimal integer $\ell$
such that there exist $\ell$ disks in $\Pi$ with radius $\varepsilon$
covering $\Pi$. If no such partition exists for some $\varepsilon > 0$ 
then the covering number is infinite. The metric entropy is defined as 
${\cal{K}}(\Pi,D,\varepsilon) = \log{{\cal{N}}(\Pi,D,\varepsilon)}$.
\end{definition}

\begin{theorem}
\label{th_B}
With probability $1-\delta$ the difference $|V(\pi) - \hat{V}(\pi)|$ is less than
$\epsilon$ simultaneously for all $\pi \in \Pi$ for the sample size:
$$
N = O \left( \frac{V_{max}}{\epsilon} 2^T(1-\underline{c})^T (\log(1/\delta) + {\cal{K}})\right)\;.
$$
\end{theorem}

\begin{proof}
Given a class of policies $\Pi$ with finite covering
number ${\cal{N}}(\Pi,D,\epsilon)$, the upper bound
$\eta = 2^T(1-\underline{c})^T$  on the likelihood ratio, and $\epsilon > 0$,
\begin{eqnarray*}
\Pr \left(\sup_{\pi \in \Pi} \left| V(\pi) - \hat V(\pi)  \right| > \epsilon \right
) \leq
8 {\cal{N}}\left(\Pi,D,\frac{\epsilon}{8}\right) \exp
\left[-\frac{1}{128}\frac{\epsilon^2 N}{ \frac {V^2_{max}(\eta-1)}{N} +
\frac{V_{max} \eta \epsilon}{8}}\right]. 
\end{eqnarray*}
Note the relationship to equation~\ref{our_Bern}. The only essential
difference is in the covering number, which takes into account the extension
from a single policy $\pi$ to the class $\Pi$. This requires the sample size
$N$ to increase accordingly to achieve the given confidence level.
The derivation is similar to uniform convergence result of
Pollard~\cite{Pollard84}(see pages 24-27), using Bernstein's inequality
instead of Hoeffding's. Solving for $N$ gives us the statement of the theorem. 
\qed
\end{proof}

Let us compare our result with a similar result for algorithm by Kearns et al.~\cite{KearnsNIPS99}:
\begin{equation}
\label{kearns}
N = O \left( \left( \frac{V_{max}}{\epsilon}\right)^2 2^{2T} VC(\Pi) 
\log(T) \left( T + \log (V_{max}/\epsilon) + \log(1/\delta) \right) \right) 
\end{equation}
both dependences are exponential in the horizon, however in our case the
dependence on $(\frac{V_{max}}{\epsilon})$ is linear rather than quadratic.
The metric entropy $\log({\cal{N}})$ takes the place of the VC dimension
$VC(\Pi)$ in terms of class complexity. This reduction in a sample size could
be explained by the fact that the former algorithm uses all trajectories 
for evaluation of any policy, while the latter uses just a subset of
trajectories. 

\begin{remark}
Let us note that a weaker bound which is remarkably similar to the equation~\ref{kearns}
could be obtained~\cite{Peshkin} using Mc-Diarmid~\cite{McDiarmid89} theorem, applicable for
a more general case:  $$N = O \left( \left
( \frac{V_{max}}{\epsilon}\right)^2 2^{2T}(1-\underline{c})^{2T} 
\left( {\cal{K}} + \log(1/\delta) \right) \right).$$ 
The proof is based on the fact that replacing one history $h_i$ in the set of
samples $h_i, i \in (1..N)$ for the estimator $\hat V_{\pi'}(\pi)$ of
equation~\ref{eq_IS}, can not change the value of the estimator by more than
$\frac{V_{max}\eta}{N}$. 
\end{remark}

\subsection{Bounding the Likelihood Ratio}

We would like to find a way to estimate a policy which minimizes sample
complexity. Remember that we are free to choose a sampling policy. We have
discussed what it means for one sampling policy $\pi$ to be optimal with
respect to another. Here we would like to consider what it means for a
sampling policy $\pi$ to be optimal with respect to a policy class $\Pi$. 
Choosing the optimal sampling policy allows us to improve bounds with regard
to exponential dependence on the horizon $T$. The idea is that if we are working 
with a policy class of a finite metric dimension, the likelihood ratio can be
upper bounded through the covering number due to the limit in combinatorial choices.
The trick is to consider sample complexity for the case of the sampling
policy being optimal in the information -theoretic sense. 

This derivation is very similar to the one of an upper bound on the minimax
regret for predicting probabilities under logarithmic
loss~\cite{CesaBianchi01,OpperHuass97}. The upper bounds on logarithmic loss
we use were first obtained by Opper and Haussler~\cite{OpperHuass97} and then
generalized by Cesa-Bianchi and Lugosi~\cite{CesaBianchi01}. The result of
Cesa-Bianchi and Lugosi is more directly related to the reinforcement
learning problem since it applies to the case of arbitrary rather than {\em
static} experts, which corresponds to learning a policy. First, we describe
the sequence prediction problem and result of Cesa-Bianchi and Lugosi, then
show how to use this result in our setup.

In a sequential prediction game $T$ symbols $h_a^T =
\mytuple{a(1),\ldots,a(T)}$ are observed sequentially. After each observation
$a(t-1)$, a learner is asked how likely it is for each value $a \in A$ to be
the {\em next} observation. The learner goal is to assign a probability
distribution $\Pr(a(t) | h_a^{t-1};\pi')$ based on the previous values. When
at the next time step $t$, the actual new observation $a(t)$ is revealed, the
learner suffers a loss $-\log(\Pr(a(t) | h_a^{t-1}; \pi')$. At the end of
the game, the learner has suffered a total loss $-\sum_{t=1}^T\log\Pr(a(t) |
h_a^{t-1}; \pi')$. Using the join distribution $\Pr(h_a^T|\pi') =
\prod_{t=1}^T \Pr(a(t) | h_a^{t-1};\pi')$ we are going to write the loss as
$-\log \Pr(h_a^T| \pi') $. When it is known that the sequences $h_a^T$ are
generated by some probability distribution $\pi$ from the class $\Pi$, we
might ask what is the worst {\em regret}: the difference in the loss between
the learner and the best expert in the target class $\Pi$ on the worst
sequence:
$$
R_T = \inf_{\pi'} \sup_{h_a^T} \left\{ -\log \Pr(h_a^T| \pi')+ \sup_{\pi
\in \Pi} \log \Pr(h_a^T| \pi) \right\}.
$$

Using the explicit solution to the minimax problem due to
Shtarkov~\cite{Shtarkov87} Cesa-Bianchi and Lugosi prove the following
theorem:
\begin{theorem}(Cesa-Bianchi and Lugosi~\cite{CesaBianchi01} theorem 3)
For any policy class $\Pi$:
$$
R_T \leq \inf_{{\epsilon}>0} \left (\log{\cal{N}}(\Pi,D,\epsilon) + 24
\int_{0}^{\epsilon} \sqrt{\log {\cal {N}}(\Pi,D,\tau)} d\tau \right),
$$
\end{theorem}
where covering number and metric entropy for the class $\Pi$, are defined
using the distance measure $D_{\infty}(\pi,\pi') \doteq \sup_{a \in A}
\left| \log \Pr(a|\pi) - \log \Pr(a|\pi') \right|. $

It is now easy to relate the problem of bounding the likelihood ratio to the
worst case regret. Intuitively, we are asking what is the worst case
likelihood ratio if we have the optimal sampling policy. Optimality means
that our sampling policy will induce action sequences with probabilities
close to the estimated policies. Remember that likelihood ratio depends only
on actions sequence $h_a$ in the history $h$ according to the Lemma~\ref{lr}.
We need to upper bound the maximum value of the ratio
$\frac{\Pr(h_a|\pi)}{\Pr(h_a|\pi')}$, which corresponds to $
\inf_{\pi'} \sup_{h_a} \left( \frac{\Pr(h_a|\pi)}{\Pr(h_a|\pi')} \right). 
$

\begin{lemma} By the definition of the maximum likelihood policy $\sup_{\pi
\in \Pi} \Pr(h_a| \pi)$ we have:
$$
\inf_{\pi'} \sup_{h_a} \left( \frac{\Pr(h_a|\pi)}{\Pr(h_a|\pi')} \right) 
\leq \inf_{\pi'} \sup_{h_a} \left\{ \frac{\sup_{\pi \in \Pi} \Pr(h_a| \pi)}{\Pr(h_a| \pi')} \right\}.
$$
\end{lemma}

Henceforth we can directly apply the results of Cesa-Bianchi and Lugosi and
get a bound of $e^{R_T}$. Note the logarithmic dependence of the bound on
$R_T$ with respect to 
the covering number ${\cal {N}}$. Moreover, since actions $a$ belong to the
finite set of actions $A$, many of the remarks of Cesa-Bianchi and
Lugosi regarding finite alphabets apply~\cite{CesaBianchi01}. In particular,
for most ``parametric'' classes---which can be parametrized by a bounded subset of
${\cal {R}}^n$ in some ``smooth'' way~\cite{CesaBianchi01}---the metric
entropy scales as follows: for some positive constants $k_1$ and $k_2$,
$$
\log {\cal{N}}(\Pi,D,\epsilon) \leq k_1 \log \frac{k_2 \sqrt{T}}{\epsilon}.
$$
For such policies the minimax regret can be bounded by 
$$
R_T \leq \frac{k_1}{2} \log T + o(\log T),
$$
which makes the likelihood ratio bound of $\eta  = O((T)^\frac{k_1}{2})$.
In this case exponential dependence on the horizon is eliminated and the
sample complexity bound becomes 
$$
N =  O \left( \frac{V_{max}}{\epsilon} T^{\frac{k_1}{2}} \left( {\cal{K}} 
+ \log 1/\delta \right) \right). 
$$

\section{Discussion and Future Work}
\label{future}

In this paper, we developed value estimators that utilize data gathered when
using one policy, to estimate the value of using another policy, resulting in
data-efficient algorithms. We considered the question of
accumulating a sufficient experience and gave PAC-style bounds. Note that for
these bounds to hold the  covering number of the class of policies $\Pi$
should be finite.

Armed with the theorem~\ref{th_B} we are ready to answer a very important
question of how to choose among several candidate policy classes. Our
reasoning here is similar to that of structural risk minimization principal
by Vapnik~\cite{Vapnik98}. The intuition is that given a very limited data,
one might prefer to work with a primitive class of hypotheses with good
confidence, rather than getting lost in a sophisticated class of hypotheses
due to low confidence. Formally, we would have the following method: given a
set of policy classes $\Pi_1,\Pi_2, \ldots$ with corresponding covering
numbers ${\cal N}_1,{\cal N}_2, \ldots$, a confidence $\delta$ and a number
of available samples $N$, compare error bounds $\epsilon_1,\epsilon_2,\ldots$
according to the theorem~\ref{th_B}. Another way to utilize the result of
theorem~\ref{th_B} is to find what is the minimal experience necessary to be
able to provide the estimate for any policy in the class with a given
confidence. This work also provides insight for a new optimization technique.
Given the value estimate, the number of samples used, and the covering number
of the policy class, one can search for optimal policies in a class using a
new cost function $\hat{V}(\pi) + \Phi({\cal N},\delta,N) \leq V(\pi)$. This
is similar in spirit to using structural risk minimization instead of
empirical risk minimization.

The capacity of the class of policies is measured by bounds on covering
numbers in our work or by VC-dimension in the work of Kearns {\it et
al.}~\cite{KearnsNIPS99}. The worst case assumptions of these bounds
often make them far too loose for practical use. An alternative would be to
use more empirical or data dependent measures of capacity, {\it e.g.} the
empirical VC dimension~\cite{VapnikLevinLeCun94} or maximal discrepancy
penalties on splits of data~\cite{BartlettBucheronLugosi00}, 
which tend to give more accurate results. 

We are currently working on extending our results for the {\em weighted
importance sampling} ({\sc wis}) estimator~\cite{Powell66,Rubinstein81} which is a
biased but consistent estimator and has a better variance for the case of
small number of samples. This can be done using martingale inequalities by
Mc-Diarmid~\cite{McDiarmid89} to parallel Bernstein's result. There is room
for employing various alternative sampling techniques, in order to
approximate the optimal sampling policy, for example one might want to
interrupt uninformative histories, which do not bring any return for a while.
Another place for algorithm sophistication is sample pruning for the case
when the set of histories gets large. A few most representative samples can
reduce the computational cost of estimation.

\subsubsection*{Acknowledgements}
Theodoros Evgeniou introduced L.P. to the field of statistical learning
theory. Leslie Kaelbling, Tommi Jaakkola, Michael Schwarz, Luis Ortiz and
anonymous reviewers gave helpful remarks.

\nocite{Haussler92}


\end{document}